\documentclass[conference]{IEEEtran}
\IEEEoverridecommandlockouts
\usepackage[noadjust]{cite}

\usepackage{xcolor}
\usepackage{amsmath,amssymb,amsfonts}
\usepackage{algorithm}
\usepackage{algpseudocode}
\usepackage{amsmath}
\usepackage{graphicx}
\usepackage{textcomp}
\usepackage{xcolor}
\def\BibTeX{{\rm B\kern-.05em{\sc i\kern-.025em b}\kern-.08em
    T\kern-.1667em\lower.7ex\hbox{E}\kern-.125emX}}
\begin{document}
\newcommand{\MY}[1]{\textcolor{blue}{#1}}

\title{A Survey of Large Language Model Agents for Question Answering\\
}

\author{\IEEEauthorblockN{Murong Yue}
\IEEEauthorblockA{\textit{Computer Science Department} \\
\textit{George Mason Univeristy}\\
Fairfax, VA-22030 \\
myue@gmu.edu}
}

\maketitle

\begin{abstract}
This paper surveys the development of large language model (LLM)-based agents for question answering (QA). Traditional agents face significant limitations, including substantial data requirements and difficulty in generalizing to new environments. LLM-based agents address these challenges by leveraging LLMs as their core reasoning engine. These agents achieve superior QA results compared to traditional QA pipelines and naive LLM QA systems by enabling interaction with external environments. We systematically review the design of LLM agents in the context of QA tasks, organizing our discussion across key stages: planning, question understanding, information retrieval, and answer generation. Additionally, this paper identifies ongoing challenges and explores future research directions to enhance the performance of LLM agent QA systems.

\end{abstract}


\section{Introduction}
The concept of autonomous agents has long been recognized in artificial intelligence research. These agents can perceive their environment and act upon it autonomously, pursuing predetermined goals \cite{franklin1996agent}. 
The rapid advancement of large language models (LLMs) has led to increased interest in LLM-based agents \cite{park2023generative,wang2024survey}. LLMs are neural networks comprising billions of parameters. Through training on vast amounts of text data, LLMs acquire a deep understanding of grammar, semantics, context, and world knowledge. This enables them to transform various natural language processing (NLP) tasks into end-to-end text generation problems, resulting in significant performance improvements across multiple domains. For LLM-based agents, we can take the perception sensor as the \textit{``eye''} and textual representative action as the \textit{``hand''}. In this scenario, the LLM serves as the \textit{``brain''} in building sophisticated agents, addressing the limitations of prior agents.
Training prior agents requires substantial sample data and high costs in expert reward design.
In contrast, LLM agents have broad world knowledge and demonstrate strong generalization capabilities to adapt to new tasks or environments. Besides, LLMs exhibit powerful reasoning skills due to their broad language understanding and comprehensive world knowledge, handling complex queries even without specific environmental training. Besides, they accept natural language input, offering flexibility, explainability, and user-friendliness.

Question answering (QA) has been a longstanding research focus in NLP and is a widely adopted application for LLM-based agents. QA aims to provide correct answers to questions based on given context or knowledge, addressing human information needs \cite{jm3}. It is worth noting that many NLP tasks can be framed in a QA format; for instance, a translation task can be posed as \textit{``Can you provide the translation of the following sentence''}. In this survey, we focus specifically on tasks where users have explicit information needs.
While LLMs can directly answer questions, they face certain limitations. Firstly, LLMs may produce hallucinations, generating imprecise or incorrect answers, particularly when nuanced, domain-specific knowledge is required. This is especially problematic in complex fields such as legal, financial, or medical decision-making \cite{cui2024chatlaw,financemath}. Secondly, LLM's inference does not interact with external environments, such as databases for retrieving up-to-date information or tools (e.g., calculators, APIs) for obtaining more accurate answers. Besides, they cannot autonomously verify the correctness of their outputs within the environment.

LLM-based agents are widely used in QA to address these issues. The key distinction between LLM-based agents and standalone LLMs in QA tasks lies in the heuristic design of multiple modules. These modules guide the LLM in performing specific actions, such as planning, and enable interaction with external environments, including databases, tools, other trained models, and humans.

This paper presents a comprehensive survey of LLM agent design for QA tasks. We begin by providing the necessary preliminary knowledge to understand the survey. We then summarize current research on LLM agents in QA, organizing our review based on each stage of the QA process into \textit{planning}, \textit{question understanding}, \textit{information retrieval}, \textit{answer generation}, and \textit{follow-up interaction}.
For each stage, we discuss the motivation for introducing this stage and explore how LLM agents are designed to enhance the performance of each stage. Additionally, we identify various challenges in this field and discuss potential future research directions.

The research covered in this survey is primarily drawn from top-tier conferences and journals in the NLP field. Key venues include the Annual Meeting of the Association for Computational Linguistics (ACL), the Conference on Empirical Methods in Natural Language Processing (EMNLP), the North American Chapter of the Association for Computational Linguistics (NAACL), the International Conference on Learning Representations (ICLR), and the Conference on Neural Information Processing Systems (NeurIPS), etc.
\section{Preliminary}
\begin{figure}[t]
    \centering
    \includegraphics[width=.8\linewidth]{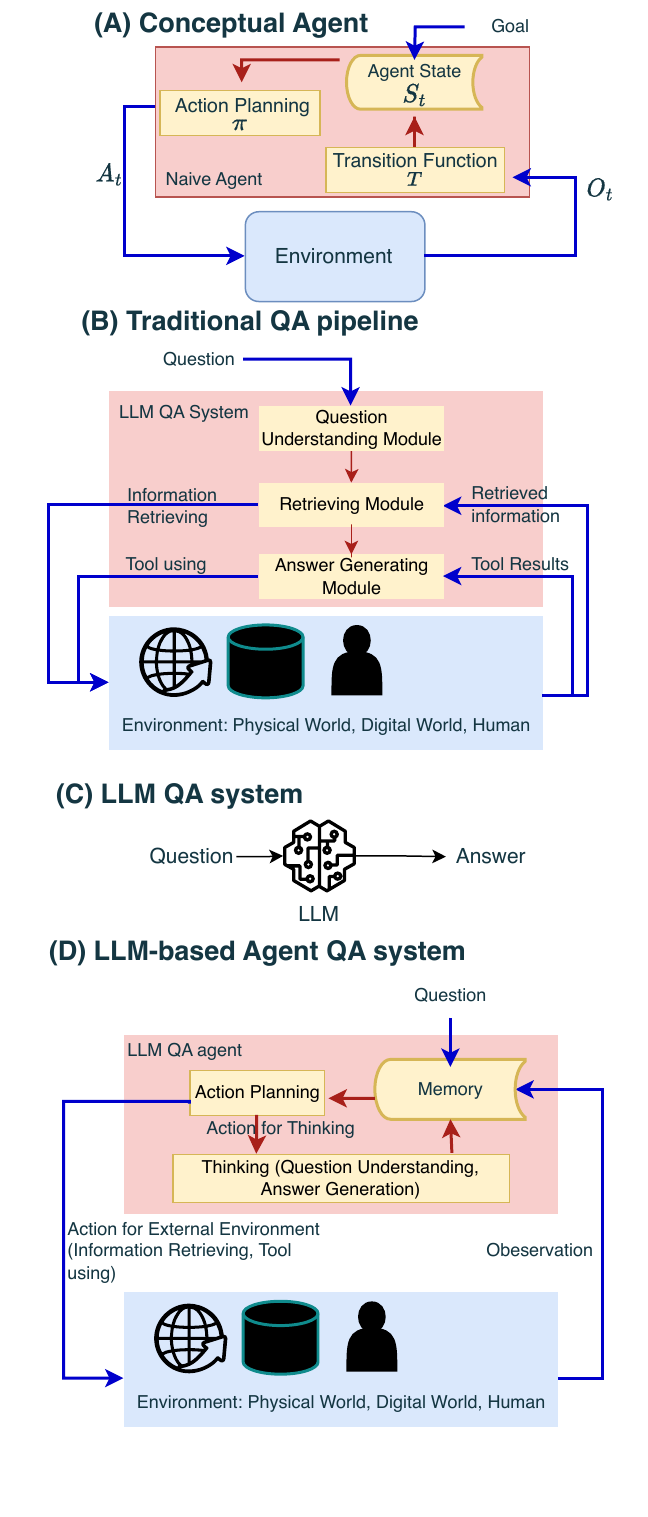}
    \vspace{-2mm}
    \caption{Overview of the naive agent, traditional QA pipeline, naive LLM QA system, and LLM-based agent QA system.}
    \label{fig:overview}
\end{figure}
\subsection{Agent}
An agent is defined as a computational entity capable of interacting with its environment to achieve specific goals.
Figure~\ref{fig:overview} (A) presents an overview of a conceptual agent. The agent typically consists of:
\begin{enumerate}
    \item State $S$: refers to a specific configuration or condition of the agent and its environment at a given point in time, such as the agent's current position, the goal of the agent, etc.
    \item Observation $O$: the information the agent perceives from its environment through sensors.
    \item Action $A$: the specific decision or behavior the agent chooses to execute in the environment.
\end{enumerate}

Let $S_t$ denote the agent's state at time $t$. The agent's planning process is represented as:
\vspace{-1mm}
\begin{equation}
A_t = \pi_p(S_t)
\end{equation}

where $\pi_p$ is the planning policy function and $A_t$ is the action at time $t$. The agent's observation $O_t$ is obtained by:
\vspace{-1mm}
\begin{equation}
O_t = E(A_t)
\end{equation}

where $E$ is the sensory processing function. The new agent state is then updated by:
\vspace{-1mm}
\begin{equation}
S_{t+1} = T(S_t, A_t, O_t)
\label{eq:transition}
\end{equation}

where $T$ is the transition function defining how an agent's state transitions from one to another.

Naive agents were designed based on heuristic or rule-based approaches, capable of handling only explicitly programmed scenarios and lacking generalization abilities \cite{nilsson1982principles}. The advent of Reinforcement Learning (RL) based agents introduced a paradigm shift, enabling agents to learn from interaction with their environment through trial and error and feedback mechanisms in the form of rewards \cite{mnih2015human,lillicrap2015continuous}.


\subsection{Question Answering}
Question Answering (QA) is a task wherein a system automatically provides answers to questions posed by users. It serves as a crucial application across multiple domains, including search engines, customer support, and education \cite{voorhees1999trec}.

Let $Q$ be the input question, $C$ be the given context, and $A = (a_1, a_2, \dots, a_T)$ be the answer sequence composed of $T$ tokens. $P(A | Q)$ is the probability of generating the answer given the question.

QA systems are generally constructed using three main approaches, categorized by the answer format:

\subsubsection{Classification-based Methods}
Given multiple possible answers $A_1, A_2, \dots, A_n$, the system is treated as a multi-class classifier:
\vspace{-1mm}
\begin{equation}
P(A_i | Q) = \text{classifier}(Q, C; \theta)
\end{equation}

where $A_i$ is the predicted answer from the set of possible answers $A_1, A_2, \dots, A_n$.

\subsubsection{Span-extraction Methods}
These methods involve identifying a span of text in a given document or context that contains the answer \cite{rajpurkar2016squad}:
\vspace{-1mm}
\begin{equation}
P_s = \text{classifier}(C_s, Q, C; \theta), \quad P_e = \text{classifier}(C_e, Q, C; \theta)
\end{equation}

where $C_s$ and $C_e$ are the tokens of the start and end positions, and $P_s$ and $P_e$ are the probabilities of the answer start and end positions in the context $C$.

\subsubsection{Text-generation Methods}
These models approach QA as a sequence generation task. Given a question, the model classifies all tokens:
\vspace{-1mm}

\begin{equation}
A = \arg\max_{A} \prod_{t=1}^{T} P(a_t \mid Q, C, a_1, a_2, \dots, a_{t-1}; \theta)
\label{eq:text_generation}
\end{equation}

where the model generates each token $a_t$ based on the question $Q$, content $C$, and previously generated tokens $a_1, a_2, \dots, a_{t-1}$.

Early efforts in text generation for QA include neural sequence-to-sequence models, where an encoder maps the input question to a latent representation, and a decoder generates the answer \cite{bahdanau2014neural}.

\subsection{Traditional QA Pipeline}
\label{sec: traditional QA}
Figure~\ref{fig:overview} (B) illustrates the traditional QA pipeline. These systems relied on decomposing the problem-solving process into several steps, executed through a fixed pipeline. The process typically begins with query understanding, which involves performing syntactic analysis and semantic understanding of the user's query~\cite{garg2019tanda}. This stage includes part-of-speech tagging, dependency parsing, intent classification, and slot extraction. Subsequently, the system employs information retrieval techniques to search for potentially relevant documents or sources of answers from a pre-established database, such as knowledge graphs, databases, or documents\cite{karpukhin2020dense}. Finally, the system selects the most probable answer from candidate answers, extracts the answer from the context, or generates the answer and presents it to the user as the final output.

The primary limitations of this pipeline approach are threefold. First, each sub-module required training a specialized model, and these models lacked world knowledge, rendering them unable to handle out-of-domain questions. Second, the pipeline was static, incapable of dynamically planning steps based on the question. Third, for open-domain questions, a vast corpus was required for retrieval, and the system relied heavily on the efficacy of the retrievers.

\subsection{LLM-based QA}
As depicted in Figure~\ref{fig:overview} (C), LLM-based QA utilizes pre-trained LLMs to comprehend and generate answers, often through fine-tuning on specific datasets or by employing few-shot prompting techniques. This approach has redefined QA paradigms.

Mainstream pre-trained LLMs can be broadly categorized into two types: Masked LLMs (e.g., BERT \cite{devlin2018bert}), which are employed in span extract QA tasks, and Autoregressive LLMs (e.g., GPT \cite{openai2023gpt4}), which are used in text generation and adhere to the formula presented in equation~\ref{eq:text_generation}. In this survey, we focus more on the latter.

In contrast to traditional QA systems, LLMs possess the capability to generate coherent and contextually relevant answers, even for questions they were not explicitly trained on. Furthermore, they can perform open-domain QA directly, leveraging their extensive knowledge encoded from large-scale pre-training \cite{petroni2019language}.

Despite their strengths, LLM-based QA models face several limitations. A significant issue is the phenomenon of hallucination, wherein LLMs may generate factually incorrect but plausible-sounding answers \cite{ji2023survey}. Additionally, LLMs are constrained by their inability to consult external databases, APIs, or other dynamic sources during inference \cite{lewis2020retrieval}. Once trained, the model's parameters are fixed, necessitating complete reliance on internalized knowledge to generate answers. 

\subsection{LLM-based QA Agent}
To address the limitations of traditional QA pipelines and naive LLM QA systems, LLM-based agents have emerged as a superior solution, as illustrated in Figure~\ref{fig:overview} (D).

The architecture of an LLM-based agent typically comprises three primary components: memory ($M$), which aggregates all information the agent possesses, including the initial question $Q$, textual results for understanding the question, and retrieved information; a planning module ($\pi_p$), which determines the next action to take by referring to the LLM with planning prompts; and an inner-thinking module ($\pi_t$), which executes inner thinking actions by referring to the LLM with thinking prompts.
The environment can be the physical world, the digital world (e.g., APIs, web browsers), or even human interactions.

LLM agents dynamically answer the question via several steps. At step $t$, $A_t$ represents the action.
The action space of an agent can be classified into external and internal actions, depending on whether interaction with the environment is involved. External actions enable the agent to engage with the environment, gather an observation $O_t$, and store it in its memory $M$. In contrast, internal actions do not involve external interaction; instead, they modify the agent's memory $M$ based purely on its intrinsic reasoning.

\begin{algorithm}[t!]
\caption{LLM Agent Question Answering}
\label{alg:llm-agent}
\begin{algorithmic}[1]
\State \textbf{Initialize:} Memory $M \gets Q$ \Comment{Initialize memory with the question $Q$}
\For{each time step $t$}
    \State $A_t \gets \pi_p(M)$ \Comment{Action planning module selects action based on memory}
    \If{$A_t$ interacts with external environment}
        \State $O_t \gets F_{Env}(A_t)$ \Comment{Environment feedback function returns observation}
        \State $M \gets M \parallel(A_t, O_t)$ \Comment{Thinking module processes action and observation to get texts of current state}
    \Else
        \State $T_t \gets F_{Think}(A_t)$ \Comment{Thinking module returns feedback}
        \State $M \gets M \parallel (A_t, T_t)$ \Comment{Thinking module processes action to get texts of current state}
    \EndIf
\EndFor
\end{algorithmic}
\end{algorithm}


The process is presented in Algorithm~\ref{alg:llm-agent}. 
The memory $M$ is initialized with the question $Q$. For time step $t$, the action $A_t$ is determined based on the planner $\pi_p$:
\vspace{-1mm}
\begin{equation}
A_t = \pi_p(M)
\end{equation}

If $A_t$ is an action for interacting with the external environment, $O_t$ is obtained by:
\vspace{-1mm}
\begin{equation}
O_t = E(A_t)
\end{equation}

where $E$ is the environment feedback function. Subsequently, the memory is updated as:
\vspace{-1mm}
\begin{equation}
M = M \parallel (A_t, O_t)
\end{equation}

where $\parallel$ denotes concatenation.

If $A_t$ is an action for thinking, the thought $T_t$ is obtained by:
\vspace{-1mm}
\begin{equation}
T_t = \pi_t(A_t)
\end{equation}

After obtaining the state, memory is updated as:
\vspace{-1mm}
\begin{equation}
M = M \parallel (A_t, T_t)
\end{equation}

The LLM-based QA agent can be considered as a specialized instance of the conceptual agent. In this framework, the planning policy network for the LLM-based agent is typically the LLM itself. The agent state from the general agent is analogous to the memory $M$ in the LLM-based agent, and the transition function $T$ in equation~\ref{eq:transition} is equivalent to the concatenation operation. 
A key distinction lies in the action space of the LLM-based agent, which encompasses not only interactions with the environment but also the activation of inner thinking processes. Using LLM as the planner and inner thinker expanded capability, endows the LLM-based agent with superior generalization and reasoning abilities compared to other agent types, such as those based on reinforcement learning.

\vspace{-1mm}
\section{Area Taxonomy} 
\begin{figure*}[t]
    \centering
    \includegraphics[width=.7\linewidth]{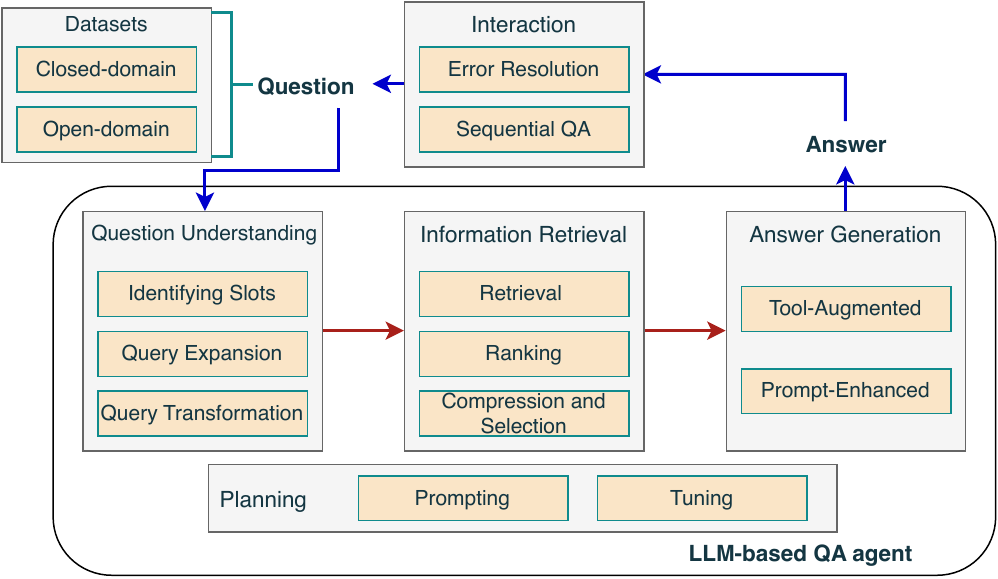}
    \caption{Area Taxonomy of LLM-based QA agent.}
    \label{fig:area_taxonomy}
\end{figure*}

As we discussed in Section~\ref{sec: traditional QA}, the question-answering process can be broken down into: \textit{question understanding}, \textit{information retrieval}, and \textit{answer generation}. Additionally,  \textit{planning} is the unique feature of the LLM agent and is the inner ability of the LLM agent where the system determines the best strategy to answer the question, such as how to understand the query, whether to retrieve information directly or infer an answer through reasoning, etc. Therefore, we follow these stages to organize our survey.
Besides, a QA system may include \textit{follow-up interaction}, which allows users to clarify their queries or ask related follow-up questions, making the interaction more dynamic and user-centric \cite{sun2022open}.
We follow this taxonomy as shown in Figure~\ref{fig:area_taxonomy} to present the cutting-edge technique used in question answering in the LLM era.

\vspace{-1mm}
\section{Taxonomy-based Survey}
\subsection{Datasets}
The rapid advancement of LLMs has significantly impacted QA tasks, prompting the creation of diverse and challenging datasets. These datasets can be broadly categorized into two main types: closed-domain/context-based QA and open-domain QA. The distinction between closed-domain/context-based QA and open-domain QA lies in whether the scope of knowledge is constrained to specific documents or not. 
In \textit{closed-domain QA}, the model’s reference is strictly limited to a predefined context, such as a passage, document, or domain-specific corpus, provided alongside the question. The model is expected to derive the answer solely from this context. External knowledge or information beyond the provided context is neither required nor considered.
In contrast, \textit{open-domain QA} involves no such constraints on the source of knowledge. The model is tasked with answering questions by retrieving relevant information from a vast, open-ended corpus, such as the internet, or from the inner knowledge acquired in the pre-training stage. This approach demands the ability to handle the queries with no specific document or passage supplied.

\paragraph{\textbf{Close-domain datasets}}
Closed-domain QA datasets are typically designed to assess a system’s ability to answer questions within specific documents.
The dataset Squad~\cite{rajpurkar2016squad} and DROP~\cite{drop} is designed to assess a system’s ability to perform comprehension over text passages. It requires models to handle complex documents and extract relevant information to answer questions. DROP also involves performing arithmetic operations or comparisons based on the extracted information.
HotpotQA~\cite{hotpotqa}, is also designed for reading comprehension but places more emphasis on multi-hop reasoning. Here, models must answer questions by analyzing information from multiple paragraphs, often requiring them to perform intermediate reasoning steps to connect evidence across different contexts. This encourages not only factual recall but also the ability to perform reason over the extracted facts.
\textit{FinQA} \cite{financemath} is a dataset specifically crafted for financial reasoning. It contains questions based on financial reports and documents, requiring models to not only extract information but also understand and reason about financial concepts, often involving numeric and arithmetic reasoning.


\paragraph{\textbf{Open-domain QA datasets}}
In contrast, \textit{open-domain QA} datasets do not provide explicit references to specific documents or passages from which the answer must be derived. Instead, these datasets cover a broader range of topics and test a wide variety of reasoning skills, often requiring the model to retrieve or infer information from vast, unstructured knowledge sources or the knowledge stored in its parameters. In the LLM era, open-domain QA datasets become more important as it's a more natural way for humans to interact with the machine.

Factual question answering in open-domain QA is exemplified by datasets such as \textit{StrategyQA} \cite{strategyqa}, which requires models to recall the information required and answer yes/no questions by employing multi-step reasoning to determine whether the claim is correct. Another dataset, \textit{ASQA} \cite{asqa}, focuses on answering ambiguous questions by generating multiple possible interpretations and corresponding answers. Meanwhile, \textit{ELI5}~ \cite{eli5} is designed for answering long-form, open-ended questions posed by non-experts, often requiring models to provide detailed explanations on a wide array of topics in a manner that is both informative and accessible to lay audiences.

When it comes to evaluating \textit{mathematical reasoning}, several challenging datasets are widely used. \textit{GSM8k} \cite{gsm8k} tests the ability to solve grade-school-level math problems, emphasizing arithmetic and problem-solving skills. For more advanced mathematical reasoning, \textit{MATH} \cite{math} and \textit{TheoremQA} \cite{chen2023theoremqa} datasets present high-school-level and university-level math problems, respectively, covering topics like algebra, calculus, geometry. etc. Additionally, \textit{Olympic Math} \cite{olympicmath} introduces a suite of competition-level problems that challenge models to solve complex and creative mathematical puzzles, often requiring deep mathematical insight.

Symbolic reasoning capabilities are tested using datasets like \textit{BBH} \cite{bbh} (Big Bench Hard), which includes a variety of difficult reasoning tasks such as pattern recognition, logic puzzles, and algorithmic reasoning. Another dataset, \textit{Folio} \cite{folio}, is specifically designed to assess the symbolic reasoning abilities of models through a range of formal logic and symbolic manipulation tasks.

For evaluating \textit{knowledge-intensive reasoning} across multiple domains, datasets such as \textit{MMLU} \cite{mmlu} are utilized. MMLU consists of questions from over 50 different fields, including history, physics, and law, and requires models to demonstrate broad knowledge and reasoning across both the humanities and sciences. Similarly, \textit{GPQA} \cite{gpqa} assesses the model's ability to answer open-ended graduate-student level questions across various domains, often requiring retrieval of specific scientific facts. \textit{WikiQA} \cite{wikiqa} focuses on open-domain question answering based on Wikipedia data, where models are tasked with retrieving relevant information from Wikipedia articles to answer a diverse set of questions.

Lastly, \textit{conditional reasoning} is explored through datasets like \textit{IFQA} \cite{ifqa}, where models must reason conditionally over conditions that may be counterfactual. The model is required to understand hypothetical or counterfactual scenarios and derive correct answers based on given conditions or changes in context. 

\subsection{Planning}
\textit{Planning} is a crucial component in autonomous systems and is key to enabling agents to take deliberate actions. It refers to the process by which an agent formulates a sequence of intermediate steps or actions that lead to the achievement of a final goal or answer.
There are two primary paradigms:
\paragraph{\textbf{Prompting-based Planning}}

In this approach, the LLM is guided by well-formulated instructions, leveraging its latent knowledge to make decisions. By prompting the LLM to consider intermediate actions and reasoning steps, the model can effectively answer the question. \textit{ReAct}~\cite{yao2022react} prompts the LLM to not only think about the next action it should take but also the content of that action. For instance, if the next step involves retrieving information, the model is asked both \textit{what} action to take (e.g., searching) and \textit{what} specific content it should search for. This approach demonstrated the potential of using prompting as a mechanism to guide the LLM's decision-making and planning capabilities. \textit{ReAct} was one of the first works to show that prompting LLMs can effectively make a plan based on the current situation and take a series of actions to better answer the question. \textit{Think on Graph}~\cite{sun2023think} focuses on planning of knowledge-graph based QA. In this approach, the LLM is prompted to make decisions about whether it should continue exploring nodes in a knowledge graph to gather additional information, or whether it has enough data to answer the question at hand. 
This strategy enables the LLM to iteratively plan its search through the graph, making dynamic decisions about when to stop gathering information and proceed with answering the question, thus improving its reasoning capabilities over structured data.
Similarly, \textit{Active Retriever}~\cite{jiang2023active} highlights the importance of continually gathering information across multiple reasoning steps.
This approach emphasizes that a single round of retrieval might not be sufficient for answering complex questions, and therefore prompts the LLM to plan multiple retrievals when necessary. The model is prompted to assess the completeness of its current information and to decide whether further retrieval is needed before attempting to answer the question. By prompting the LLM to actively plan repeated rounds of information gathering, \textit{Active Retriever} ensures that the model remains flexible and can adjust its strategy as it progresses through the task.
\textit{Agentverse}~\cite{chen2023agentverse} extends this idea of planning by prompting LLMs to decide which expert models or agents to involve in answering the question during the decision-making process. In this approach, the LLM is tasked with choosing between various specialized models or retrieval systems, based on the nature of the question. For example, the LLM might decide that a specific expert model is better suited to retrieve legal or financial information. 

Collectively, these works demonstrate the effectiveness of prompting LLMs to make the plan.
However, despite the success of these methods, they heavily rely on the careful design of heuristic prompts. The effectiveness of the LLM’s planning ability depends significantly on how well the instructions are formulated and how similar the \textit{in-context demonstrations}, i.e., the examples provided within the prompt, are provided. This dependence on prompt design poses a challenge for generalization. Because each prompt and set of demonstrations is often tailored to a specific task or domain, the LLM may struggle to transfer its planning abilities to new, unseen contexts. 

\paragraph{\textbf{Tuning-based Planning}}
In these approaches, the LLM can learn from incorrect action trajectories and refine its strategies through trial and error, improving its ability to solve tasks autonomously.
One approach is \textit{FireAct}~\cite{chen2023fireact}, which fine-tunes LLMs using action trajectories generated from muti-hop QA tasks. This method capitalizes on the ReAct-style framework, creating multiple potential action trajectories for solving a task. During training, GPT-4 is prompted to generate these action trajectories. The correct action trajectories are then collected and used to fine-tune the planner, allowing it to learn from past attempts and gradually improve its decision-making processes. 
The \textit{Learning from Failure} approach~\cite{wang2024learning} suggests that using only the correct trajectories, while discarding failed attempts, leads to significant wastage of valuable data. This approach recognizes that the comparison between successful and failed trajectories can offer crucial insights for fine-tuning the planner. Instead of ignoring failed trajectories, \textit{Learning from Failure} proposes incorporating them into the training process. Specifically, the method includes positive examples with a standard prompt and negative examples with a special prompt that indicates the case was incorrect. By learning from both success and failure, the model can better understand why certain actions lead to failure and how they can be avoided in future tasks. During inference, only the normal prompt is provided, but the planner has already been fine-tuned to understand the difference between successful and unsuccessful trajectories. This method enhances the model’s ability to generalize from both types of experiences, offering a more holistic approach to learning from trial and error.
While many prior works are limited to a narrow range of tasks, \textit{AgentGen}~\cite{hu2024agentgen} aims to synthesize planning paths for a much more diverse set of tasks, each conditioned on specific environments. \textit{AgentGen} first constructs multiple seed environments. The LLM is then prompted to alter these environments, for example by \textit{``adding more constraints''}. Once the modified environment is generated, the LLM proceeds to collect trajectories based on the new environment. This allows the model to explore various ways of handling different types of environments, significantly expanding its planning capabilities beyond the narrow focus of prior methods.

The primary advantage of these tuning-based planning approaches is their ability to learn from vast amounts of training data, including both successful and failed attempts. 
However, these methods also have notable limitations. One major challenge is that they rely heavily on \textit{searching} through multiple potential trajectories to identify the best planning path. This reliance on search-based methods can limit the scalability of the approach.
Additionally, the fine-tuning process itself, while effective at improving task-specific performance, can negatively impact the model’s ability to generalize to new, unseen tasks, posing a trade-off between task-specific optimization and broader generalization.

\subsection{Question Understanding}
The question-understanding process requires extracting and comprehending information from the user’s query and making it easier for the machine to understand. Users' questions can be ambiguous or complex, so multiple techniques are used to help the machine process the question to make the answer easier.
Traditionally, separate models were trained specifically for tasks like slot tagging and intent understanding~\cite{chen2019ICSF}. 
As LLMs show the inherent ability to handle complex linguistic structures, the research made them useful for performing question understanding without the need for task-specific models.
\paragraph{\textbf{Identifying Slots}}
\textit{Slot identification} focuses on recognizing specific entities, variables, or attributes within a query and categorizing them based on predefined types, such as names, dates, locations, or specialized terms. This process serves as a bridge between unstructured natural language input and structured data representations, enabling systems to map user queries into a structured format that can be processed.
An example of the slot identifying of the LLM agent is the \textit{ChatLaw}~\cite{cui2024chatlaw}, a legal consultation system that utilizes LLMs to identify and cluster legal entities from consultation questions. 
For example, when a user asks a legal question such as \textit{``What are the penalties for breach of contract in California?''}, the system extracts key entities like \textit{``California''} as location. 
It serves as a basic step of the question understanding.

\paragraph{\textbf{Query Expansion}}
\textit{Query expansion} enhances the retrieval of relevant information by augmenting the user's original query with additional terms or phrases. These additional terms can include synonyms, related concepts, or more specific details that are inferred from the context of the search or question~\cite{carpineto2012}. 
In many cases, users' initial queries may lack the precision needed to return the most relevant results. 
By expanding the query with terms that are semantically related to the original input, query expansion helps to mitigate these issues. 
For example, if a user queries \textit{``car insurance claims,''} a query expansion process might add terms like \textit{``vehicle insurance,''} \textit{``auto claims,''} or \textit{``accident report''} to improve the retrieval of documents that might not explicitly match the user's initial input but are still relevant to the topic~\cite{robertson2004}.

An approach utilizing LLMs for query expansion is \textit{HyQE}~\cite{gao-etal-2023-hyqe}, which prompts LLMs to generate multiple hypothetical documents that act as expansions for the original query. The key insight behind \textit{HyQE} is that hypothetical documents, generated by LLMs, have a higher probability of containing the necessary keywords that are likely to be present in relevant answer documents. 
Another notable approach is \textit{Query2CoT}~\cite{jagerman2023querycot}. It first breaks down complex queries into step-by-step sub-questions. The LLM is instructed to identify the key keywords associated with each sub-question. By decomposing the query in this way, \textit{Query2CoT} enables the retrieval system to focus on different components of the query in isolation, improving precision in identifying relevant documents. Another innovative query expansion method is \textit{Step-back reformulation}~\cite{zheng2023takestepback}, which reformulates complex reasoning questions into higher-level concept questions. 
This method focuses on simplifying reasoning-intensive queries by stepping back from the detailed question and focusing on a broader conceptual understanding. 
For example, given the specific question \textit{``Estella Leopold went to which school between August 1954 and November 1954?''}, a step-back reformulation might simplify this into the higher-level question \textit{``What was Estella Leopold's education history?''}. By expanding the query in this way, the system broadens its scope of inquiry, allowing for the retrieval of more general information that might still provide the answer.

\paragraph{\textbf{Query Reformulation}}
Another effective approach to handling ambiguity or vagueness in user queries is \textit{query reformulation}. This technique involves rephrasing or simplifying. 
A prominent method for query reformulation is \textit{Rephrase and Response}~\cite{deng2023rephrase}, which designs specific prompts to instruct LLM to rewrite questions for improved clarity. In this approach, the LLM is prompted to rephrase the original query in a more structured or precise manner, helping to clarify ambiguities and improve the alignment between the query and the retrieved information. 
In addition to prompt-based methods, other approaches have explored fine-tuning LLMs specifically for the task of query rewriting~\cite{ma2023query, Peng2023Taobao}. The authors introduce methods where LLMs are fine-tuned to rewrite queries into multiple alternative versions. The system first generates multiple reformulations of the original query and then evaluates which of these reformulated queries leads to the most accurate answer in a downstream pipeline. In these methods, the correctness of the final answer acts as a reward signal to guide the selection of the best query rewrite. Once the best-performing query is identified, the system uses techniques like \textit{Direct Preference Optimization (DPO)}~\cite{rafailov2024direct} to fine-tune the LLM, training it to generate improved query rewrites in future interactions.

\subsection{Information Retrieval}
The \textit{information retrieval} component in LLM-based QA agents is pivotal in answering knowledge-intensive questions for extracting relevant knowledge from a vast corpus or external knowledge sources. Information retrieval refers to the process of identifying and ranking documents, passages, or snippets that may contain the necessary information to answer a given question. This process typically follows a retrieving and ranking paradigm:
\paragraph{\textbf{Retrieval}} 
Retrieval involves fetching candidate documents or passages using sparse or dense retrieval techniques. Sparse methods, like BM25 \cite{jm3}, use term frequency-inverse document frequency (TF-IDF) based algorithms. However, sparse methods often struggle with semantic mismatches where the query and document might have a similar meaning but use different terms.

Dense retrieval methods learn vector representations for queries and documents, embedding them in a shared semantic space. Once the query $q$ and document $d$ are embedded into vectors, we compute the similarity between the two using a distance metric and retrieve top-K documents.
The objective of dense retrieval can be formulated using a contrastive learning objective: \begin{equation} \mathcal{L} = \log \frac{e^{sim(q_i, d_i^+)}}{e^{sim(q_i, d_i^+)} + \sum_{j=1}^{N} e^{sim(q_i, d_j^-)}} \end{equation} where $d_i^+$ is the positive sample, $d_j^-$ are negative samples, and $N$ is the number of negative samples \cite{gao2024modular}.

\paragraph{\textbf{Ranking}}
Although the retrieved documents can help the LLM agent in solving questions with external knowledge, sometimes it may hurt the performance while providing irrelevant documents when answering~\cite{yoran2023making}. Besides, the number of the input tokens of LLM is limited. Therefore, it is important to get the most relevant information after retrieval.
Ranking typically follows retrieval, where an additional model, such as the LLM itself or a tuned LLM, assigns scores to the retrieved candidates, refining their relevance. For a given query $q$ and retrieved documents ${d_1, d_2, ..., d_k}$, the ranking model produces a relevance score $r(q, d_i)$ for each document. The documents are then ordered by their scores: \begin{equation} r(q, d_i) = \text{Rank Model}(q, d_i) \end{equation} 

Recent research, including studies by Ma et al.~\cite{ma2023large} and Zhuang et al.~\cite{zhuang2023open}, highlights that while LLMs may not excel as document retrievers, they demonstrate remarkable capabilities when employed as re-rankers of previously retrieved documents. 
The evaluation compared the performance of LLM-based rankers against traditional rankers. LLMs show significant strengths in re-ranking retrieved documents. 
Since LLMs possess a powerful ability to comprehend and analyze the deeper semantic meanings within documents, they can assign relevance scores that align more closely with the user’s intent or query.

An example of an LLM-based system that leverages this re-ranking ability is \textit{Haystack}~\cite{ranker}. In \textit{Haystack}, LLMs are prompted to assess the relevance of a set of documents to a given query. These scores are then used to re-rank the retrieved documents, ensuring that the most semantically relevant documents appear higher in the list.
Another approach is \textit{Self-RAG}~\cite{asai2023self}. In \textit{Self-RAG}, a critical LLM is fine-tuned to evaluate not only the relevance of retrieved documents but also their usefulness in contributing to the final answer. One of the key challenges in retrieval-augmented generation systems is determining when to trust the LLM's internal knowledge versus relying on external documents. The critical LLM in \textit{Self-RAG} addresses this issue by distinguishing between documents that provide novel information and those that simply repeat knowledge that the LLM already possesses. 
For example, the critical LLM will be in favor of documents that add new or complementary information. 

\paragraph{\textbf{Compression and Selection}}
Compression and selection techniques summarize long documents and select the most relevant passages. The key insight behind this technique is that not all portions of a document contribute equally to the final output. These techniques aim to preserve the essential content while reducing the length of the document. The compression model maps a document $D$ to a compressed representation $C(D)$: \begin{equation} C(D) = \text{Compression Model}(D) \end{equation} 

\textit{LLMLingua}~\cite{jiang2023longllmlingua} introduces a novel coarse-to-fine, step-by-step compression methodology designed specifically to handle long prompts. 
\textit{LLMLingua} first applies a coarse-grained compression step to reduce the input size by removing extraneous or low-importance information. 
This is followed by a fine-grained process that further refines the compressed input, ensuring that essential semantic content is preserved. 
By progressively compressing the prompt in this manner, \textit{LLMLingua} manages to reduce the input length significantly while retaining the necessary information to maintain the model’s performance. 
\textit{RRecomp}~\cite{xu2023recomp} fine-tunes LLMs as both extractive and abstractive compressors. The extractive compression phase involves identifying and retaining the most important sentences or phrases, which ensures that critical details are preserved. Following the extractive phase, \textit{RRecomp} applies an abstractive compression step, where the remaining content is summarized or rephrased into a more concise form. This hybrid model of compression allows RRecomp to balance detail retention and brevity.


\subsection{Answer Generation}
Answer generation synthesizes relevant information to produce a response to the given query. Several methods can enhance this process in LLM-based agents:
\paragraph{\textbf{Tool-augmented Generation}}
This approach allows LLMs to interact with external tools, such as calculators or code interpreters, to augment their reasoning capabilities. 
The \textit{Program-of-Thought} (PoT) approach~\cite{chen2022pot,gao2023pal} focuses on using LLMs to generate executable Python code as part of the reasoning process. Instead of relying solely on the LLM to generate answers directly, PoT leverages the code interpreter to get the final answer. The LLM generates Python code as intermediate steps, which can be executed in a code interpreter, and the results are returned as the final answer. 
The advantage of this approach is that it allows LLMs to handle problems that require both complex reasoning and precise computation.

\textit{HuggingGPT}~\cite{shen2024hugginggpt} and \textit{OpenAGI}~\cite{ge2024openagi} extend this concept further by integrating domain-specific external models as tools for LLMs. These systems prompt the LLM to recognize tasks that require specialized knowledge beyond its internal capacity and then invoke external models to handle those tasks. For instance, in an image-related task, HuggingGPT can call upon a pre-trained image segmentation model to analyze an image, extract relevant features, or make predictions, which the LLM then incorporates into its reasoning process to produce a final answer.
Similarly, OpenAGI acts as an orchestrator that connects LLMs to various task-specific models—whether for processing audio, images, or other modalities—allowing the LLM to integrate multimodal information into its responses. 
This approach enables the LLM to work alongside specialized AI models, thereby enhancing its ability to answer questions or solve problems that extend beyond the textual or conceptual knowledge encoded in its pre-trained weights. 

The \textit{Binding} approach~\cite{cheng2022binding,gao2024efficient} takes this integration of external tools one step further by using a hybrid system of natural language generation and targeted computational tools. In Binding, the LLM is first prompted to generate a preliminary solution to a problem in the form of a natural language skeleton. However, instead of producing a complete solution, the LLM intentionally leaves certain parts of the answer blank as placeholders for information that will be calculated or filled in by external tools. For example, in solving a mathematical word problem, the LLM might generate the structure of the solution in natural language, outlining the steps required to solve the problem, but leaving placeholders for specific numerical calculations. These masked tokens are then filled in using external tools like a calculator, which can compute the precise values needed. 
Binding improves the overall reliability of LLM-generated answers by ensuring that numerical or factual inaccuracies are minimized through the use of dedicated external computational systems.


\paragraph{\textbf{Prompt-Enhanced Generation}}
Various prompting techniques can improve answer generation. 

The \textit{Chain-of-Thought} (CoT) approach~\cite{wei2022chain} is to prompt LLMs to generate step-by-step intermediate reasoning steps when answering complex questions, rather than providing direct answers in a single pass. This structured reasoning process allows the model to break down the solving into smaller, manageable steps. 
One of the advancements in CoT is the introduction of \textit{question decomposition} techniques. Zhou et al.~\cite{zhou2023leasttomostpromptingenablescomplex} propose the \textit{Least-to-Most} prompting strategy, where complex questions are explicitly decomposed into a series of simpler sub-questions that can be tackled in sequence. 
Another notable refinement is the concept of \textit{self-verification}~\cite{chen2023teaching}. Self-verification leverages the model's own reasoning to verify the correctness of its intermediate steps. After generating an initial answer, the LLM is prompted to re-evaluate its reasoning by checking the final answer for logical consistency and factual accuracy. 

Recent approaches optimize the prompting with the training data. \textit{Self-Discovery}~\cite{zhou2024self-discover} aims to automate the discovery of the composition of optimal reasoning strategies. In Self-Discovery, the LLM is trained to explore different ways of approaching the problem, experimenting with diverse prompts and reasoning strategies. Over time, the model learns which strategies lead to more successful outcomes and adjusts its behavior accordingly. 
The \textit{PromptAgent}~\cite{wang2023promptagent} builds on these ideas by integrating automatic prompt optimization into the reasoning process. PromptAgent treats prompt construction as an adaptive, multi-step process where the LLM itself plays a central role in refining and improving the prompt structure. PromptAgent continuously updates the prompt based on the LLM’s performance during question answering. 


\subsection{Follow-up Interaction}
Follow-up interaction maintains engagement and ensures clarity during conversations. Unlike traditional QA systems, LLM agents engage in multi-turn conversations, refining answers based on user feedback \cite{wang2023mint}.
\paragraph{\textbf{Error Resolution}}
Follow-up interaction allows models to clarify misunderstandings or request additional details. Schick et al. \cite{schick2022peer} explore using human feedback to revise creative writing, while Yan et al. \cite{yan2023learning} apply human feedback to refine semantic parsing results.
\paragraph{\textbf{Sequential Question Answering}}
This involves maintaining context across multiple question-answering processes \cite{iyyer2017search}. LLMs can engage in multi-turn conversations after instruction tuning and reinforcement learning from human feedback \cite{ouyang2022training}. ChatQA \cite{liu2024chatqa} further enhances this ability through context-enhanced instruction tuning.
\section{Open Problem}
As LLM QA agents grow more powerful, tasks that once seemed challenging, such as generating coherent multi-turn conversations or solving complex math problems, have now become more manageable. However, it is still far from perfect, and new challenges also arise.

\subsection{Challenging and Practical Benchmarking}
As LLM agents show great performance improvement, the challenge of benchmarking them becomes increasingly complex. It includes multiple aspects:

\paragraph{Fine-grained Answer Generation Process Evaluation} Evaluating free-form answers remains a major challenge for existing LLM-based QA system benchmarks. Many benchmarks still rely on multiple-choice formats or coarse-grained metrics like ROGUE~\cite{lin2004rouge} and BERTScore~\cite{zhang2019bertscore}, which focus solely on the final answer. Such evaluations overlook the reasoning process that leads to the answer, limiting their granularity. With the increased adoption of techniques like CoT~\cite{wei2022chain}, future benchmarks need to incorporate more fine-grained evaluation mechanisms that assess not only the final answer but also the underlying reasoning process, ensuring both the correctness and coherence of the thought process~\cite{mondorf2024beyond}.

\paragraph{Hard-to-judge Question Evaluation} Some questions are inherently difficult to assess, unlike games such as Go, where the criteria for success are clear. These questions can be categorized into two types. First, those with an objectively correct answer but where verifying correctness is difficult, such as complex math problems that may take even experts several hours or days to assess. Second, open-ended questions, such as \textit{``How to improve a QA system?''} where no absolute answer exists. Developing benchmarks that effectively evaluate LLM agents' performance on such questions is crucial to understanding their path toward achieving human-level intelligence.

\paragraph{Up-to-date Evaluation} A pressing challenge is maintaining the relevance and fairness of benchmarks over time. As LLMs continue to learn from vast and constantly evolving internet data, preventing data leakage and ensuring fair comparisons become essential. First, it is necessary to protect the integrity of the dataset's content, such as test questions and answers. A possible solution would be to develop a benchmark that dynamically updates itself~\cite{ying2024automatingdatasetupdatesreliable}. Second, benchmarks should protect the structure and format of datasets, as using synthetic data or hand-labeled examples to manipulate leaderboards~\cite{dubey2024llama} can lead to overfitting. Currently, there are few effective mechanisms to prevent this, making the development of more credible and time-resilient benchmarks a valuable research direction.

\subsection{Hallucination and Calibration}
One of the most pressing issues with current LLMs in QA is their tendency to hallucinate, i.e. generating false or fabricated information while maintaining confidence that their output is accurate. Unlike humans, who can often gauge their confidence in an answer, LLMs lack well-calibrated mechanisms for judging the correctness of their outputs. Addressing this issue requires improving the model’s calibration ability, enabling it to better predict when it may be wrong and communicate that uncertainty. There are several possible avenues for tackling this challenge:

\paragraph{Integrating External Tools/Knowledge} A promising research direction involves integrating external tools or knowledge databases that can help LLMs assess the confidence levels of their responses~\cite{semnani2023wikichat}. However, one challenge is that LLMs can be overly influenced by external information without adequately assessing its reliability, unlike humans who are more attuned to the credibility of information sources. Therefore, enabling LLMs to autonomously analyze and weigh the reliability of different sources is an important direction for reducing hallucinations.

\paragraph{Improving Calibration Ability in Training LLM} Another essential approach is to improve the internal calibration mechanisms of LLMs during training. Techniques such as uncertainty-aware training~\cite{yang2023improving} aim to fine-tune models to predict their uncertainty during training. However, these methods have not yet shown significant improvements compared to traditional logit-based methods or ensemble approaches like voting-based strategies~\cite{lin2023generating}. Thus, developing more effective training methods that enhance LLMs' inherent calibration abilities remains an open problem.

\paragraph{LLM Intrinsic Representation} Humans typically experience awareness when they are unsure of an answer, raising the question: can LLMs develop a similar form of self-awareness? And if so, how can it be represented? This area of research might involve identifying specific neural circuits within the model that correspond to overconfidence or uncertainty and then adjusting these inner states accordingly. By detecting and modulating these intrinsic representations, researchers could help LLMs become more aware of their own reliability, allowing them to provide more trustworthy answers~\cite{orgad2024llmsIntrinsicRepresentation}. However, this is an emerging field, and the underlying mechanisms are still not well understood.

\subsection{Reasoning Ability Improvement}
Enhancing the reasoning capabilities of LLM agents in QA remains an important and challenging area of exploration. Reasoning ability is crucial in QA because it not only determines the correctness of answers but also ensures that the process leading to the conclusion is logical, interpretable, and reliable. Therefore, improving reasoning skills is key to making LLMs effective in real-world problem-solving scenarios.

\paragraph{Exploration to Enhance Reasoning}
One promising approach to improving reasoning is to explore multiple reasoning paths for training questions or synthesized data~\cite{chan2024scaling}. Studies have shown that searching through various trials and training LLMs on successful paths can significantly enhance their reasoning capabilities~\cite{AlphaGeometryTrinh2024}. However, current search methods often rely on some domain-specific signals, such as Lean language or algebraic representations in geometric problems, which limit the generalization to general domains. In other areas, reward models~\cite{zhang2024rest} or LLM self-evaluation strategies~\cite{tian2024toward} are used to score the reasoning process. Therefore, developing more reliable scoring mechanisms for reasoning path exploration is a critical step toward improving LLM reasoning across diverse fields.

\paragraph{Improving Reasoning from Memory}
Another key area is enabling LLMs to learn from memory, which allows them to quickly adapt their reasoning to new environments. Unlike humans, who learn from past experiences, LLMs treat each interaction as a new session, often repeating the same mistakes. Enhancing LLMs' ability to retain and learn from past interactions is essential. This goes beyond simply providing interaction history at inference time. LLMs need to be able to extract foundational knowledge from past experiences and apply it to new scenarios, thereby improving their adaptability and reasoning effectiveness.

\paragraph{Enhancing Logical Chains with Causal Reasoning}
Integrating causal reasoning can significantly improve the rigor and coherence of LLM-generated logical chains. Typically, LLMs generate responses based on statistical correlations, which may not accurately reflect the causal logic required for certain problems. By incorporating causal reasoning frameworks, LLMs can better identify and apply causal relationships within problems, leading to more robust logical reasoning. This approach not only improves the accuracy of answers but also enhances their interpretability. Training LLMs to recognize and utilize causal models, especially in domains where causal structures are inherent, will help generate more precise and insightful responses.

\subsection{Autonomous Tool Selection and Creation}
Humans possess the ability to select appropriate tools for various tasks, often summarizing their experiences to create new tools tailored for specific purposes. In the context of QA, our humans instinctively determine when to seek external assistance, such as searching the web, consulting a database, or asking an expert. In contrast, current LLMs lack the innate ability to actively choose tools or external resources. They do not inherently plan the steps required to solve a problem or determine when to involve external systems, presenting a significant challenge in developing more autonomous, intelligent agents capable of utilizing the right resources at the right time. Besides, they cannot find the pattern of questions and create tools by themselves to solve some repeated questions.

To address this limitation, future research should focus on enabling LLMs to develop a form of tool-use and tool-creation planning. This would allow the model to dynamically decide when to leverage the tools and when to create new tools. Such capabilities would bring LLMs closer to human-like problem-solving, enhancing their effectiveness in open-ended tasks and collaborative environments.

\subsection{LLMs in Building Document Indexing}
Another key challenge is the role of LLMs in improving information retrieval (IR). 
Given that LLMs have demonstrated the ability to grasp the semantic meanings of natural language, their integration into the retrieval process shows great promise. Currently, LLMs are employed for tasks such as question expansion/formulation or ranking to enhance IR performance.
From our perspective, incorporating LLMs into document indexing represents a novel promising direction for future research. 
Indexing involves converting documents into vector representations using an embedding model, allowing document vectors to be retrieved based on the similarity to query vectors. A primary challenge is the cost associated with using LLMs to index millions or billions of documents. However, as smaller LLMs continue to improve in capability, integrating LLMs into the indexing process may become a viable and impactful research avenue.

\section{Conclusion}
The rapid development of LLM agents has significantly enhanced question-answering systems. This survey recalls the development of the agent and the QA systems and then defines the concept of LLM agent QA systems. We break down the answering process into multiple sub-tasks, demonstrating how cutting-edge methods are leveraged in improving the LLM agent QA system. 
Finally, we highlight notable challenges and identify promising avenues of research that could elevate LLM-based agents into even more powerful ones.

\bibliographystyle{IEEEtran}
\bibliography{ref}

\begin{thebibliography}{10}
\providecommand{\url}[1]{#1}
\csname url@samestyle\endcsname
\providecommand{\newblock}{\relax}
\providecommand{\bibinfo}[2]{#2}
\providecommand{\BIBentrySTDinterwordspacing}{\spaceskip=0pt\relax}
\providecommand{\BIBentryALTinterwordstretchfactor}{4}
\providecommand{\BIBentryALTinterwordspacing}{\spaceskip=\fontdimen2\font plus
\BIBentryALTinterwordstretchfactor\fontdimen3\font minus \fontdimen4\font\relax}
\providecommand{\BIBforeignlanguage}[2]{{%
\expandafter\ifx\csname l@#1\endcsname\relax
\typeout{** WARNING: IEEEtran.bst: No hyphenation pattern has been}%
\typeout{** loaded for the language `#1'. Using the pattern for}%
\typeout{** the default language instead.}%
\else
\language=\csname l@#1\endcsname
\fi
#2}}
\providecommand{\BIBdecl}{\relax}
\BIBdecl

\bibitem{franklin1996agent}
S.~Franklin and A.~Graesser, ``Is it an agent, or just a program?: A taxonomy for autonomous agents,'' in \emph{International workshop on agent theories, architectures, and languages}.\hskip 1em plus 0.5em minus 0.4em\relax Springer, 1996, pp. 21--35.

\bibitem{park2023generative}
J.~S. Park, J.~O'Brien, C.~J. Cai, M.~R. Morris, P.~Liang, and M.~S. Bernstein, ``Generative agents: Interactive simulacra of human behavior,'' in \emph{Proceedings of the 36th annual acm symposium on user interface software and technology}, 2023, pp. 1--22.

\bibitem{wang2024survey}
L.~Wang, C.~Ma, X.~Feng, Z.~Zhang, H.~Yang, J.~Zhang, Z.~Chen, J.~Tang, X.~Chen, Y.~Lin \emph{et~al.}, ``A survey on large language model based autonomous agents,'' \emph{Frontiers of Computer Science}, vol.~18, no.~6, p. 186345, 2024.

\bibitem{jm3}
D.~Jurafsky and J.~H. Martin, \emph{Speech and Language Processing: An Introduction to Natural Language Processing, Computational Linguistics, and Speech Recognition with Language Models}, 2024.

\bibitem{cui2024chatlaw}
J.~Cui, M.~Ning, Z.~Li, B.~Chen, Y.~Yan, H.~Li, B.~Ling, Y.~Tian, and L.~Yuan, ``Chatlaw: A multi-agent collaborative legal assistant with knowledge graph enhanced mixture-of-experts large language model,'' 2024.

\bibitem{financemath}
W.~Tao, H.~Zhu, K.~Tan, J.~Wang, Y.~Liang, H.~Jiang, P.~Yuan, and Y.~Lan, ``Finqa: A training-free dynamic knowledge graph question answering system in finance with llm-based revision,'' in \emph{Joint European Conference on Machine Learning and Knowledge Discovery in Databases}.\hskip 1em plus 0.5em minus 0.4em\relax Springer, 2024, pp. 418--423.

\bibitem{nilsson1982principles}
N.~J. Nilsson, \emph{Principles of artificial intelligence}.\hskip 1em plus 0.5em minus 0.4em\relax Springer, 1982.

\bibitem{mnih2015human}
V.~Mnih \emph{et~al.}, ``Human-level control through deep reinforcement learning,'' \emph{Nature}, vol. 518, no. 7540, pp. 529--533, 2015.

\bibitem{lillicrap2015continuous}
T.~P. Lillicrap \emph{et~al.}, ``Continuous control with deep reinforcement learning,'' \emph{arXiv preprint arXiv:1509.02971}, 2015.

\bibitem{voorhees1999trec}
E.~M. Voorhees and D.~M. Tice, ``The trec-8 question answering track report,'' in \emph{Text retrieval conference TREC}.\hskip 1em plus 0.5em minus 0.4em\relax Citeseer, 1999.

\bibitem{rajpurkar2016squad}
P.~Rajpurkar \emph{et~al.}, ``Squad: 100,000+ questions for machine comprehension of text,'' in \emph{Proceedings of the 2016 Conference on Empirical Methods in Natural Language Processing}, 2016, pp. 2383--2392.

\bibitem{bahdanau2014neural}
D.~Bahdanau \emph{et~al.}, ``Neural machine translation by jointly learning to align and translate,'' \emph{arXiv preprint arXiv:1409.0473}, 2014.

\bibitem{garg2019tanda}
S.~Garg \emph{et~al.}, ``Tanda: Transfer and adapt pre-trained transformer models for answer sentence selection,'' in \emph{Proceedings of the 57th Annual Meeting of the Association for Computational Linguistics}, 2019, pp. 5488--5494.

\bibitem{karpukhin2020dense}
V.~Karpukhin \emph{et~al.}, ``Dense passage retrieval for open-domain question answering,'' in \emph{Proceedings of the 2020 Conference on Empirical Methods in Natural Language Processing (EMNLP)}, 2020, pp. 6769--6781.

\bibitem{devlin2018bert}
J.~Devlin \emph{et~al.}, ``Bert: Pre-training of deep bidirectional transformers for language understanding,'' \emph{arXiv preprint arXiv:1810.04805}, 2018.

\bibitem{openai2023gpt4}
OpenAI, ``Gpt-4 technical report,'' \emph{arXiv preprint arXiv:2303.08774}, 2023.

\bibitem{petroni2019language}
F.~Petroni \emph{et~al.}, ``Language models as knowledge bases?'' \emph{arXiv preprint arXiv:1909.01066}, 2019.

\bibitem{ji2023survey}
Z.~Ji \emph{et~al.}, ``A survey of hallucination in natural language generation,'' \emph{ACM Computing Surveys (CSUR)}, vol.~55, no.~12, pp. 1--38, 2023.

\bibitem{lewis2020retrieval}
P.~Lewis \emph{et~al.}, ``Retrieval-augmented generation for knowledge-intensive nlp tasks,'' in \emph{Advances in Neural Information Processing Systems}, 2020, pp. 9459--9474.

\bibitem{sun2022open}
F.~Zhu, W.~Lei, C.~Wang, J.~Zheng, S.~Poria, and T.-S. Chua, ``Retrieving and reading: A comprehensive survey on open-domain question answering,'' \emph{arXiv preprint arXiv:2101.00774}, 2021.

\bibitem{drop}
D.~Dua, Y.~Wang, P.~Dasigi, G.~Stanovsky, S.~Singh, and M.~Gardner, ``Drop: A reading comprehension benchmark requiring discrete reasoning over paragraphs,'' \emph{arXiv preprint arXiv:1903.00161}, 2019.

\bibitem{hotpotqa}
Z.~Yang, P.~Qi, S.~Zhang, Y.~Bengio, W.~W. Cohen, R.~Salakhutdinov, and C.~D. Manning, ``Hotpotqa: A dataset for diverse, explainable multi-hop question answering,'' in \emph{Proceedings of the 2018 Conference on Empirical Methods in Natural Language Processing}, 2018, pp. 2369--2380.

\bibitem{strategyqa}
M.~Geva, D.~Khashabi, E.~Segal, T.~Khot, D.~Roth, and J.~Berant, ``Did aristotle use a laptop? a question answering benchmark with implicit reasoning strategies,'' \emph{Transactions of the Association for Computational Linguistics}, vol.~9, pp. 346--361, 2021.

\bibitem{asqa}
I.~Stelmakh, Y.~Luan, B.~Dhingra, and M.-W. Chang, ``Asqa: Factoid questions meet long-form answers,'' \emph{arXiv preprint arXiv:2204.06092}, 2022.

\bibitem{eli5}
A.~Fan, P.~Lewis, and Y.~Dauphin, ``Eli5: Long-form question answering,'' \emph{arXiv preprint arXiv:1907.09190}, 2019.

\bibitem{gsm8k}
K.~Cobbe, V.~Kosaraju, M.~Bavarian, M.~Chen, H.~Jun, L.~Kaiser, M.~Plappert, J.~Tworek, J.~Hilton, R.~Nakano, C.~Hesse, and J.~Schulman, ``Training verifiers to solve math word problems,'' \emph{arXiv preprint arXiv:2110.14168}, 2021.

\bibitem{math}
D.~Hendrycks, C.~Burns, S.~Kadavath, A.~Arora, S.~Basart, E.~Tang, D.~Song, and J.~Steinhardt, ``Measuring mathematical problem solving with the math dataset,'' \emph{arXiv preprint arXiv:2103.03874}, 2021.

\bibitem{chen2023theoremqa}
W.~Chen, M.~Yin, M.~Ku, P.~Lu, Y.~Wan, X.~Ma, J.~Xu, X.~Wang, and T.~Xia, ``Theoremqa: A theorem-driven question answering dataset,'' in \emph{Proceedings of the 2023 Conference on Empirical Methods in Natural Language Processing}, 2023, pp. 7889--7901.

\bibitem{olympicmath}
C.~He, R.~Luo, Y.~Bai, S.~Hu, Z.~L. Thai, J.~Shen, J.~Hu, X.~Han, Y.~Huang, Y.~Zhang, J.~Liu, L.~Qi, Z.~Liu, and M.~Sun, ``Olympiadbench: A challenging benchmark for promoting agi with olympiad-level bilingual multimodal scientific problems,'' 2024.

\bibitem{bbh}
M.~Suzgun, N.~Scales, N.~Sch{\"a}rli, S.~Gehrmann, Y.~Tay, H.~W. Chung, A.~Chowdhery, Q.~V. Le, E.~H. Chi, D.~Zhou, , and J.~Wei, ``Challenging big-bench tasks and whether chain-of-thought can solve them,'' \emph{arXiv preprint arXiv:2210.09261}, 2022.

\bibitem{folio}
\BIBentryALTinterwordspacing
S.~Han, H.~Schoelkopf, Y.~Zhao, Z.~Qi, M.~Riddell, L.~Benson, L.~Sun, E.~Zubova, Y.~Qiao, M.~Burtell, D.~Peng, J.~Fan, Y.~Liu, B.~Wong, M.~Sailor, A.~Ni, L.~Nan, J.~Kasai, T.~Yu, R.~Zhang, S.~Joty, A.~R. Fabbri, W.~Kryscinski, X.~V. Lin, C.~Xiong, and D.~Radev, ``Folio: Natural language reasoning with first-order logic,'' \emph{arXiv preprint arXiv:2209.00840}, 2022. [Online]. Available: \url{https://arxiv.org/abs/2209.00840}
\BIBentrySTDinterwordspacing

\bibitem{mmlu}
D.~Hendrycks, C.~Burns, S.~Basart, A.~Zou, M.~Mazeika, D.~Song, and J.~Steinhardt, ``Measuring massive multitask language understanding,'' \emph{arXiv preprint arXiv:2009.03300}, 2020.

\bibitem{gpqa}
\BIBentryALTinterwordspacing
D.~Rein, B.~L. Hou, A.~C. Stickland, J.~Petty, R.~Y. Pang, J.~Dirani, J.~Michael, and S.~R. Bowman, ``{GPQA}: A graduate-level google-proof q\&a benchmark,'' in \emph{First Conference on Language Modeling}, 2024. [Online]. Available: \url{https://openreview.net/forum?id=Ti67584b98}
\BIBentrySTDinterwordspacing

\bibitem{wikiqa}
Y.~Yang, W.-t. Yih, and C.~Meek, ``Wikiqa: A challenge dataset for open-domain question answering using wikipedia,'' \emph{arXiv preprint arXiv:1412.7808}, 2015.

\bibitem{ifqa}
W.~Yu, M.~Jiang, P.~Clark, and A.~Sabharwal, ``Ifqa: A dataset for open-domain question answering under counterfactual presuppositions,'' \emph{arXiv preprint arXiv:2305.14010}, 2023.

\bibitem{yao2022react}
S.~Yao, J.~Zhao, D.~Yu, N.~Du, I.~Shafran, K.~Narasimhan, and Y.~Cao, ``React: Synergizing reasoning and acting in language models,'' \emph{arXiv preprint arXiv:2210.03629}, 2022.

\bibitem{sun2023think}
J.~Sun, C.~Xu, L.~Tang, S.~Wang, C.~Lin, Y.~Gong, H.-Y. Shum, and J.~Guo, ``Think-on-graph: Deep and responsible reasoning of large language model with knowledge graph,'' \emph{arXiv preprint arXiv:2307.07697}, 2023.

\bibitem{jiang2023active}
Z.~Jiang, F.~F. Xu, L.~Gao, Z.~Sun, Q.~Liu, J.~Dwivedi-Yu, Y.~Yang, J.~Callan, and G.~Neubig, ``Active retrieval augmented generation,'' \emph{arXiv preprint arXiv:2305.06983}, 2023.

\bibitem{chen2023agentverse}
W.~Chen, Y.~Su, J.~Zuo, C.~Yang, C.~Yuan, C.-M. Chan, H.~Yu, Y.~Lu, Y.-H. Hung, C.~Qian \emph{et~al.}, ``Agentverse: Facilitating multi-agent collaboration and exploring emergent behaviors,'' in \emph{The Twelfth International Conference on Learning Representations}, 2023.

\bibitem{chen2023fireact}
B.~Chen, C.~Shu, E.~Shareghi, N.~Collier, K.~Narasimhan, and S.~Yao, ``Fireact: Toward language agent fine-tuning,'' \emph{arXiv preprint arXiv:2310.05915}, 2023.

\bibitem{wang2024learning}
R.~Wang, H.~Li, X.~Han, Y.~Zhang, and T.~Baldwin, ``Learning from failure: Integrating negative examples when fine-tuning large language models as agents,'' \emph{arXiv preprint arXiv:2402.11651}, 2024.

\bibitem{hu2024agentgen}
M.~Hu, P.~Zhao, C.~Xu, Q.~Sun, J.~Lou, Q.~Lin, P.~Luo, S.~Rajmohan, and D.~Zhang, ``Agentgen: Enhancing planning abilities for large language model based agent via environment and task generation,'' \emph{arXiv preprint arXiv:2408.00764}, 2024.

\bibitem{chen2019ICSF}
Q.~Chen, Z.~Zhuo, and W.~Wang, ``Bert for joint intent classification and slot filling,'' \emph{arXiv preprint arXiv:1902.10909}, 2019.

\bibitem{carpineto2012}
C.~Carpineto and G.~Romano, ``A survey of automatic query expansion in information retrieval,'' \emph{ACM Computing Surveys (CSUR)}, vol.~44, no.~1, pp. 1--50, 2012.

\bibitem{robertson2004}
S.~Robertson, ``Understanding inverse document frequency: on theoretical arguments for idf,'' \emph{Journal of documentation}, vol.~60, no.~5, pp. 503--520, 2004.

\bibitem{gao-etal-2023-hyqe}
\BIBentryALTinterwordspacing
L.~Gao, X.~Ma, J.~Lin, and J.~Callan, ``Precise zero-shot dense retrieval without relevance labels,'' in \emph{Proceedings of the 61st Annual Meeting of the Association for Computational Linguistics (Volume 1: Long Papers)}, A.~Rogers, J.~Boyd-Graber, and N.~Okazaki, Eds.\hskip 1em plus 0.5em minus 0.4em\relax Toronto, Canada: Association for Computational Linguistics, Jul. 2023, pp. 1762--1777. [Online]. Available: \url{https://aclanthology.org/2023.acl-long.99}
\BIBentrySTDinterwordspacing

\bibitem{jagerman2023querycot}
R.~Jagerman, H.~Zhuang, Z.~Qin, X.~Wang, and M.~Bendersky, ``Query expansion by prompting large language models,'' \emph{arXiv preprint arXiv:2305.03653}, 2023.

\bibitem{zheng2023takestepback}
H.~S. Zheng, S.~Mishra, X.~Chen, H.-T. Cheng, E.~H. Chi, Q.~V. Le, and D.~Zhou, ``Take a step back: Evoking reasoning via abstraction in large language models,'' \emph{arXiv preprint arXiv:2310.06117}, 2023.

\bibitem{deng2023rephrase}
Y.~Deng, W.~Zhang, Z.~Chen, and Q.~Gu, ``Rephrase and respond: Let large language models ask better questions for themselves,'' \emph{arXiv preprint arXiv:2311.04205}, 2023.

\bibitem{ma2023query}
\BIBentryALTinterwordspacing
X.~Ma, Y.~Gong, P.~He, hai zhao, and N.~Duan, ``Query rewriting in retrieval-augmented large language models,'' in \emph{The 2023 Conference on Empirical Methods in Natural Language Processing}, 2023. [Online]. Available: \url{https://openreview.net/forum?id=gXq1cwkUZc}
\BIBentrySTDinterwordspacing

\bibitem{Peng2023Taobao}
\BIBentryALTinterwordspacing
W.~Peng, G.~Li, Y.~Jiang, Z.~Wang, D.~Ou, X.~Zeng, Tongxu, and E.~Chen, ``Large language model based long-tail query rewriting in taobao search,'' \emph{Companion Proceedings of the ACM on Web Conference 2024}, 2023. [Online]. Available: \url{https://api.semanticscholar.org/CorpusID:265042961}
\BIBentrySTDinterwordspacing

\bibitem{rafailov2024direct}
R.~Rafailov, A.~Sharma, E.~Mitchell, C.~D. Manning, S.~Ermon, and C.~Finn, ``Direct preference optimization: Your language model is secretly a reward model,'' \emph{Advances in Neural Information Processing Systems}, vol.~36, 2024.

\bibitem{gao2024modular}
Y.~Gao, Y.~Xiong, M.~Wang, and H.~Wang, ``Modular rag: Transforming rag systems into lego-like reconfigurable frameworks,'' \emph{arXiv preprint arXiv:2407.21059}, 2024.

\bibitem{yoran2023making}
O.~Yoran, T.~Wolfson, O.~Ram, and J.~Berant, ``Making retrieval-augmented language models robust to irrelevant context,'' \emph{arXiv preprint arXiv:2310.01558}, 2023.

\bibitem{ma2023large}
Y.~Ma, Y.~Cao, Y.~Hong, and A.~Sun, ``Large language model is not a good few-shot information extractor, but a good reranker for hard samples!'' \emph{arXiv preprint arXiv:2303.08559}, 2023.

\bibitem{zhuang2023open}
S.~Zhuang, B.~Liu, B.~Koopman, and G.~Zuccon, ``Open-source large language models are strong zero-shot query likelihood models for document ranking,'' \emph{arXiv preprint arXiv:2310.13243}, 2023.

\bibitem{ranker}
V.~Blagojevi, ``Enhancing rag pipelines in haystack: Introducing diversityranker and lostinthemiddleranker,'' 2023.

\bibitem{asai2023self}
A.~Asai, Z.~Wu, Y.~Wang, A.~Sil, and H.~Hajishirzi, ``Self-rag: Learning to retrieve, generate, and critique through self-reflection,'' \emph{arXiv preprint arXiv:2310.11511}, 2023.

\bibitem{jiang2023longllmlingua}
H.~Jiang, Q.~Wu, X.~Luo, D.~Li, C.-Y. Lin, Y.~Yang, and L.~Qiu, ``Longllmlingua: Accelerating and enhancing llms in long context scenarios via prompt compression,'' \emph{arXiv preprint arXiv:2310.06839}, 2023.

\bibitem{xu2023recomp}
F.~Xu, W.~Shi, and E.~Choi, ``Recomp: Improving retrieval-augmented lms with compression and selective augmentation,'' \emph{arXiv preprint arXiv:2310.04408}, 2023.

\bibitem{chen2022pot}
W.~Chen, X.~Ma, X.~Wang, and W.~W. Cohen, ``Program of thoughts prompting: Disentangling computation from reasoning for numerical reasoning tasks,'' \emph{arXiv preprint arXiv:2211.12588}, 2022.

\bibitem{gao2023pal}
L.~Gao, A.~Madaan, S.~Zhou, U.~Alon, P.~Liu, Y.~Yang, J.~Callan, and G.~Neubig, ``Pal: Program-aided language models,'' in \emph{International Conference on Machine Learning}.\hskip 1em plus 0.5em minus 0.4em\relax PMLR, 2023, pp. 10\,764--10\,799.

\bibitem{shen2024hugginggpt}
Y.~Shen, K.~Song, X.~Tan, D.~Li, W.~Lu, and Y.~Zhuang, ``Hugginggpt: Solving ai tasks with chatgpt and its friends in hugging face,'' \emph{Advances in Neural Information Processing Systems}, vol.~36, 2024.

\bibitem{ge2024openagi}
Y.~Ge, W.~Hua, K.~Mei, J.~Tan, S.~Xu, Z.~Li, Y.~Zhang \emph{et~al.}, ``Openagi: When llm meets domain experts,'' \emph{Advances in Neural Information Processing Systems}, vol.~36, 2024.

\bibitem{cheng2022binding}
Z.~Cheng, T.~Xie, P.~Shi, C.~Li, R.~Nadkarni, Y.~Hu, C.~Xiong, D.~Radev, M.~Ostendorf, L.~Zettlemoyer \emph{et~al.}, ``Binding language models in symbolic languages,'' \emph{arXiv preprint arXiv:2210.02875}, 2022.

\bibitem{gao2024efficient}
S.~Gao, J.~Dwivedi-Yu, P.~Yu, X.~E. Tan, R.~Pasunuru, O.~Golovneva, K.~Sinha, A.~Celikyilmaz, A.~Bosselut, and T.~Wang, ``Efficient tool use with chain-of-abstraction reasoning,'' \emph{arXiv preprint arXiv:2401.17464}, 2024.

\bibitem{wei2022chain}
J.~Wei, X.~Wang, D.~Schuurmans, M.~Bosma, F.~Xia, E.~Chi, Q.~V. Le, D.~Zhou \emph{et~al.}, ``Chain-of-thought prompting elicits reasoning in large language models,'' \emph{Advances in neural information processing systems}, vol.~35, pp. 24\,824--24\,837, 2022.

\bibitem{zhou2023leasttomostpromptingenablescomplex}
\BIBentryALTinterwordspacing
D.~Zhou, N.~Schärli, L.~Hou, J.~Wei, N.~Scales, X.~Wang, D.~Schuurmans, C.~Cui, O.~Bousquet, Q.~Le, and E.~Chi, ``Least-to-most prompting enables complex reasoning in large language models,'' 2023. [Online]. Available: \url{https://arxiv.org/abs/2205.10625}
\BIBentrySTDinterwordspacing

\bibitem{chen2023teaching}
X.~Chen, M.~Lin, N.~Sch{\"a}rli, and D.~Zhou, ``Teaching large language models to self-debug,'' \emph{arXiv preprint arXiv:2304.05128}, 2023.

\bibitem{zhou2024self-discover}
P.~Zhou, J.~Pujara, X.~Ren, X.~Chen, H.-T. Cheng, Q.~V. Le, E.~H. Chi, D.~Zhou, S.~Mishra, and H.~S. Zheng, ``Self-discover: Large language models self-compose reasoning structures,'' \emph{arXiv preprint arXiv:2402.03620}, 2024.

\bibitem{wang2023promptagent}
X.~Wang, C.~Li, Z.~Wang, F.~Bai, H.~Luo, J.~Zhang, N.~Jojic, E.~P. Xing, and Z.~Hu, ``Promptagent: Strategic planning with language models enables expert-level prompt optimization,'' \emph{arXiv preprint arXiv:2310.16427}, 2023.

\bibitem{wang2023mint}
X.~Wang, Z.~Wang, J.~Liu, Y.~Chen, L.~Yuan, H.~Peng, and H.~Ji, ``Mint: Evaluating llms in multi-turn interaction with tools and language feedback,'' \emph{arXiv preprint arXiv:2309.10691}, 2023.

\bibitem{schick2022peer}
T.~Schick, J.~Dwivedi-Yu, Z.~Jiang, F.~Petroni, P.~Lewis, G.~Izacard, Q.~You, C.~Nalmpantis, E.~Grave, and S.~Riedel, ``Peer: A collaborative language model,'' \emph{arXiv preprint arXiv:2208.11663}, 2022.

\bibitem{yan2023learning}
H.~Yan, S.~Srivastava, Y.~Tai, S.~I. Wang, W.-t. Yih, and Z.~Yao, ``Learning to simulate natural language feedback for interactive semantic parsing,'' \emph{arXiv preprint arXiv:2305.08195}, 2023.

\bibitem{iyyer2017search}
M.~Iyyer, W.-t. Yih, and M.-W. Chang, ``Search-based neural structured learning for sequential question answering,'' in \emph{Proceedings of the 55th Annual Meeting of the Association for Computational Linguistics (Volume 1: Long Papers)}, 2017, pp. 1821--1831.

\bibitem{ouyang2022training}
L.~Ouyang, J.~Wu, X.~Jiang, D.~Almeida, C.~Wainwright, P.~Mishkin, C.~Zhang, S.~Agarwal, K.~Slama, A.~Ray \emph{et~al.}, ``Training language models to follow instructions with human feedback,'' \emph{Advances in neural information processing systems}, vol.~35, pp. 27\,730--27\,744, 2022.

\bibitem{liu2024chatqa}
Z.~Liu, W.~Ping, R.~Roy, P.~Xu, M.~Shoeybi, and B.~Catanzaro, ``Chatqa: Building gpt-4 level conversational qa models,'' \emph{arXiv preprint arXiv:2401.10225}, 2024.

\bibitem{lin2004rouge}
C.-Y. Lin, ``Rouge: A package for automatic evaluation of summaries,'' in \emph{Text summarization branches out}, 2004, pp. 74--81.

\bibitem{zhang2019bertscore}
T.~Zhang, V.~Kishore, F.~Wu, K.~Q. Weinberger, and Y.~Artzi, ``Bertscore: Evaluating text generation with bert,'' \emph{arXiv preprint arXiv:1904.09675}, 2019.

\bibitem{mondorf2024beyond}
P.~Mondorf and B.~Plank, ``Beyond accuracy: Evaluating the reasoning behavior of large language models--a survey,'' \emph{arXiv preprint arXiv:2404.01869}, 2024.

\bibitem{ying2024automatingdatasetupdatesreliable}
\BIBentryALTinterwordspacing
J.~Ying, Y.~Cao, Y.~Bai, Q.~Sun, B.~Wang, W.~Tang, Z.~Ding, Y.~Yang, X.~Huang, and S.~Yan, ``Automating dataset updates towards reliable and timely evaluation of large language models,'' 2024. [Online]. Available: \url{https://arxiv.org/abs/2402.11894}
\BIBentrySTDinterwordspacing

\bibitem{dubey2024llama}
A.~Dubey, A.~Jauhri, A.~Pandey, A.~Kadian, A.~Al-Dahle, A.~Letman, A.~Mathur, A.~Schelten, A.~Yang, A.~Fan \emph{et~al.}, ``The llama 3 herd of models,'' \emph{arXiv preprint arXiv:2407.21783}, 2024.

\bibitem{semnani2023wikichat}
S.~J. Semnani, V.~Z. Yao, H.~C. Zhang, and M.~S. Lam, ``Wikichat: Stopping the hallucination of large language model chatbots by few-shot grounding on wikipedia,'' \emph{arXiv preprint arXiv:2305.14292}, 2023.

\bibitem{yang2023improving}
Y.~Yang, H.~Li, Y.~Wang, and Y.~Wang, ``Improving the reliability of large language models by leveraging uncertainty-aware in-context learning,'' \emph{arXiv preprint arXiv:2310.04782}, 2023.

\bibitem{lin2023generating}
Z.~Lin, S.~Trivedi, and J.~Sun, ``Generating with confidence: Uncertainty quantification for black-box large language models,'' \emph{arXiv preprint arXiv:2305.19187}, 2023.

\bibitem{orgad2024llmsIntrinsicRepresentation}
H.~Orgad, M.~Toker, Z.~Gekhman, R.~Reichart, I.~Szpektor, H.~Kotek, and Y.~Belinkov, ``Llms know more than they show: On the intrinsic representation of llm hallucinations,'' \emph{arXiv preprint arXiv:2410.02707}, 2024.

\bibitem{chan2024scaling}
X.~Chan, X.~Wang, D.~Yu, H.~Mi, and D.~Yu, ``Scaling synthetic data creation with 1,000,000,000 personas,'' \emph{arXiv preprint arXiv:2406.20094}, 2024.

\bibitem{AlphaGeometryTrinh2024}
T.~Trinh, Y.~Wu, Q.~Le, H.~He, and T.~Luong, ``Solving olympiad geometry without human demonstrations,'' \emph{Nature}, 2024.

\bibitem{zhang2024rest}
D.~Zhang, S.~Zhoubian, Y.~Yue, Y.~Dong, and J.~Tang, ``Rest-mcts*: Llm self-training via process reward guided tree search,'' \emph{arXiv preprint arXiv:2406.03816}, 2024.

\bibitem{tian2024toward}
Y.~Tian, B.~Peng, L.~Song, L.~Jin, D.~Yu, H.~Mi, and D.~Yu, ``Toward self-improvement of llms via imagination, searching, and criticizing,'' \emph{arXiv preprint arXiv:2404.12253}, 2024.

\end{thebibliography}

\end{document}